\documentclass[sigconf]{acmart}
\AtBeginDocument{%
  }

\copyrightyear{2026}
\acmYear{2026}
\setcopyright{cc}
\setcctype{by}
\acmConference[MM '26]{Proceedings of the 34th ACM International Conference on Multimedia}
  {November 10--14, 2026}{Rio de Janeiro, Brazil}
\acmBooktitle{Proceedings of the 34th ACM International Conference on Multimedia
  (MM '26), November 10--14, 2026, Rio de Janeiro, Brazil}
\acmDOI{10.1145/3767308.3835540}
\acmISBN{979-8-4007-2213-4/2026/11}
\usepackage{amsmath}
\usepackage{booktabs}
\usepackage{multirow}
\usepackage{graphicx}
\usepackage{tabularx}
\usepackage{subcaption}
\usepackage{xcolor}
\usepackage{colortbl}
\usepackage{algorithm}
\usepackage{algorithmic}
\usepackage{enumitem}
\microtypesetup{expansion=false}
\newcommand{\Rcontext}{R_{\text{context}}}
\newcommand{\Rtool}{R_{\text{tool}}}

\newcommand{\Dcomp}{\Delta_{\mathrm{comp}}}
\newcolumntype{Y}{>{\centering\arraybackslash}X}

\newif\ifshowrevisions
\showrevisionsfalse
\definecolor{RevisionBlue}{HTML}{005EA8}
\definecolor{RevisionRed}{HTML}{B42318}
\definecolor{RevisionYellow}{HTML}{FFF3BF}
\newenvironment{revblock}[1]
  {\par\begingroup\noindent\ifshowrevisions\color{RevisionBlue}\textbf{\textcolor{RevisionRed}{[REV-#1]}}\ \fi}
  {\par\endgroup}
\newcommand{\revadd}[2]{%
  \ifshowrevisions
    \textcolor{RevisionRed}{\textbf{[REV-#1]}}\ \textcolor{RevisionBlue}{#2}%
  \else
    #2%
  \fi}

\begin{document}

\title{MIRAGE: How Conversation State Shapes Historical Evidence Use in Multimodal Personal Agents}

\author{Yu Liu}
\authornote{Equal contribution.}
\affiliation{%
  \institution{Institute of Information Engineering, Chinese Academy of Sciences}
  \city{Beijing}
  \country{China}}
\affiliation{%
  \institution{School of Cyber Security, University of Chinese Academy of Sciences}
  \city{Beijing}
  \country{China}}
\email{liuyu@iie.ac.cn}

\author{Wenxiao Zhang}
\authornotemark[1]
\affiliation{%
  \institution{Department of Computer Science and Software Engineering, \\The University of Western Australia}
  \city{Perth}
  \country{Australia}}
\email{wenxiao.zhang@research.uwa.edu.au}

\author{Cheng Hu}
\affiliation{%
  \institution{Institute of Information Engineering, Chinese Academy of Sciences}
  \city{Beijing}
  \country{China}}
\affiliation{%
  \institution{School of Cyber Security, University of Chinese Academy of Sciences}
  \city{Beijing}
  \country{China}}
\email{hucheng@iie.ac.cn}

\author{Cong Cao}
\affiliation{%
  \institution{Institute of Information Engineering, Chinese Academy of Sciences}
  \city{Beijing}
  \country{China}}
\email{caocong@iie.ac.cn}

\author{Fangfang Yuan}
\correspondingauthor
\affiliation{%
  \institution{Institute of Information Engineering, Chinese Academy of Sciences}
  \city{Beijing}
  \country{China}}
\email{yuanfangfang@iie.ac.cn}

\author{Xinyu Wang}
\affiliation{%
  \institution{Department of Computer Science and Software Engineering, \\The University of Western Australia}
  \city{Perth}
  \country{Australia}}
\email{xinyu.wang@research.uwa.edu.au}

\author{Jin B. Hong}
\correspondingauthor
\affiliation{%
  \institution{Department of Computer Science and Software Engineering, \\The University of Western Australia}
  \city{Perth}
  \country{Australia}}
\email{jin.hong@uwa.edu.au}

\author{Yanbing Liu}
\correspondingauthor
\affiliation{%
  \institution{Institute of Information Engineering, Chinese Academy of Sciences}
  \city{Beijing}
  \country{China}}
\affiliation{%
  \institution{School of Cyber Security, University of Chinese Academy of Sciences}
  \city{Beijing}
  \country{China}}
\email{liuyanbing@iie.ac.cn}

\renewcommand{\shortauthors}{Liu et al.}

%%% ------- ABSTRACT -------
\begin{abstract}
Multimodal large language model (MLLM) agents are increasingly used as personal assistants for long-running tasks. Their utility depends on continuity: agents must retrieve and use earlier evidence across dialogue, files, and workspace state. However, agents can generate plausible answers even when access to that history has degraded, causing outcome-only evaluation to overestimate true evidence use.
We present \textbf{MIRAGE} (\textbf{M}ultimodal \textbf{I}nteraction \textbf{R}etrieval, \textbf{A}ttribution, and \textbf{G}rounding \textbf{E}valuation), a controlled study of historical evidence use under conversation-state variation in multimodal personal agents. MIRAGE holds evidence objects, questions, and scoring fixed while varying only conversation state, and evaluates whether an agent can determine answerability, recover the correct source, and answer from it.
Across seven frontier and open-weight multimodal backbones, we find that: \textbf{1)}~pre-compaction depth and post-compaction continuation form distinct, non-monotonic failure regimes rather than a single degradation curve; \textbf{2)}~open-weight models rely heavily on context continuity and are reluctant to spontaneously switch to tool-mediated retrieval when provenance fails; and \textbf{3)}~retrieval pressure improves source attribution in deep pre-compaction states for tool-compliant models, but consistently regresses after compaction, where stored evidence has already degraded. These findings show that historical evidence use should be evaluated under state variation, rather than inferred from outcome-only correctness.
\end{abstract}




\maketitle

\begin{figure}[t]
  \centering
  \includegraphics[width=\columnwidth]{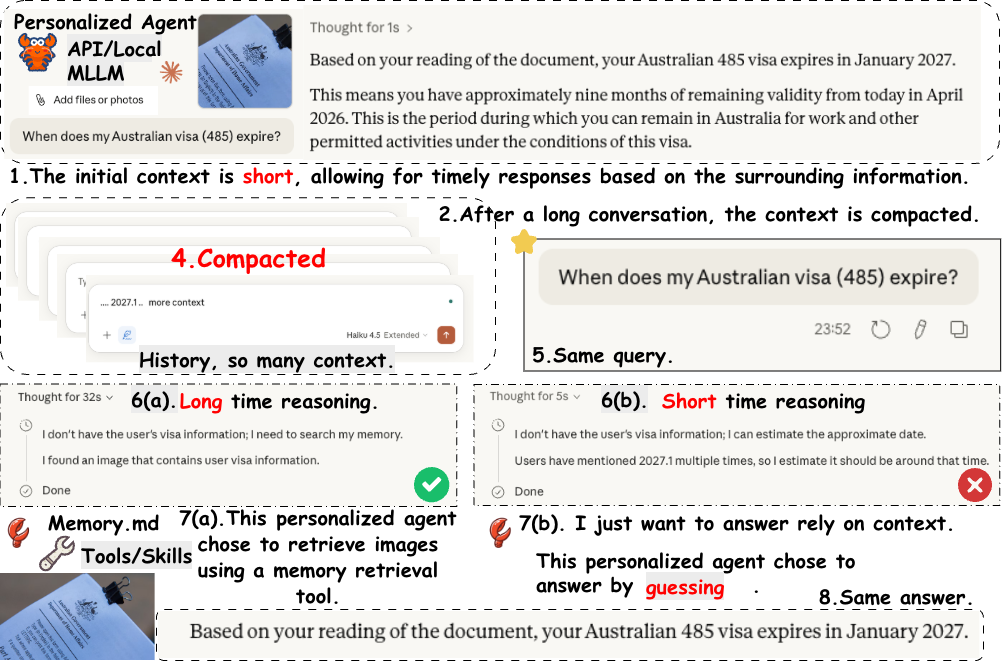}
  \caption{Same query, same answer, different grounding. After context compaction~(4), one reasoning path retrieves the source document via a memory tool~(7a); the other guesses from residual context~(7b). Both produce the same final answer~(8), making the grounding failure invisible to outcome-only evaluation.}
  \label{fig:motivation}
\end{figure}

%%% ============================================================
\section{Introduction}\label{sec:intro}
%%% ============================================================

%%% -- Paragraph 1: Setting + blind spot --
Multimodal large language model (MLLM) agents increasingly serve as long-lived personal assistants that manage ongoing tasks, revisit prior materials, and maintain continuity across dialogue, documents, images, and workspace files.
Their practical value depends on recovering and using user-specific evidence encountered earlier.
As conversations grow or cross compaction boundaries, that evidence may no longer be directly visible, causing the agent to rely on residual context, summaries, or generic priors and produce plausible but weakly grounded answers.
Because existing evaluations largely score final answers or task completion, an agent can appear to remember without grounding its response in the correct historical evidence.
Figure~\ref{fig:motivation} illustrates this blind spot: the same query about a historical document can yield the same plausible answer through either tool-mediated evidence retrieval or residual-context guessing after compaction, which outcome-only evaluation scores identically.\footnote{Code: \url{https://github.com/prisma-research/MIRAGE}}

%%% -- Paragraph 3: Prior work and the gap --
Prior work illuminates parts of this problem, but not the full setting we study.
Recent memory benchmarks study long-horizon conversational recall and multi-session memory use~\cite{maharana2024locomo,wu2025longmemeval,memgallery2026,mei2026atmbench,long2026m3agent}.
In parallel, work on retrieval faithfulness shows that tool use does not guarantee genuine evidence reliance~\cite{hallucination_survey2025,ma2025proofofuse}.
However, because existing work rarely isolates conversation state as the experimental variable, it remains unclear how evidence use changes when the same evidence and questions are tested at different context depths before and after compaction.
%%% -- Paragraph 4: Our solution --

To address this gap, we introduce \textbf{MIRAGE} (\textbf{M}ultimodal \textbf{I}nteraction \textbf{R}etrieval, \textbf{A}ttribution, and \textbf{G}rounding \textbf{E}valuation), an empirical study of historical evidence use under conversation-state variation.
We focus on multimodal materials encountered during prior interaction, including user interface views, statistical graphics, and profile views, along with their derived workspace representations.
This type of evidence is especially vulnerable to state-conditioned failure because it passes through lossy representation shifts across states (e.g., chart image $\to$ extracted table $\to$ compacted summary), making the grounding chain longer and more fragile than for plain text.
When asked about this material later, the agent must determine answerability, recover the correct source, and answer from that source rather than from a plausible guess.
Instead of proposing a large-scale leaderboard benchmark, MIRAGE deliberately keeps the planted evidence set, trunk conversation, question bank, and scoring protocol fixed so that conversation state is the only experimental variable, analogous to a factorial experiment with a fixed stimulus set. The current instantiation evaluates the same evidence-grounded questions across four states spanning shallow pre-compaction context, deeper same-session context, and post-compaction continuation, using seven frontier and open-weight multimodal backbones under two query conditions.
Our experiments follow a \emph{what--why--how} progression: (RQ1)~what failure patterns emerge as conversation state changes, (RQ2)~what retrieval mechanisms produce these patterns, and (RQ3)~when retrieval pressure can mitigate provenance failure across different conversation states. Our contributions are as follows:
\begin{enumerate}[leftmargin=*,nosep]
\item We introduce MIRAGE, a state-conditioned evaluation protocol that holds evidence, questions, and scoring fixed while varying conversation state, decomposing each response into answerability, source attribution, and evidence-grounded answering.

\item We show that pre-compaction depth and post-compaction continuation are distinct failure regimes with non-monotonic degradation, and that open-weight models rely heavily on context continuity and are reluctant to spontaneously adopt tool-mediated retrieval even when provenance has failed.

\item We demonstrate that a retrieval-pressure extension recovers source attribution at deep pre-compaction states for tool-compliant models, but consistently regresses after compaction, establishing that compaction-induced evidence degradation cannot be addressed by retrieval alone.
\end{enumerate}

%%% ============================================================
\section{Related Work}\label{sec:related}
%%% ============================================================

\subsection{Historical Evidence in Long-Horizon Agent Systems}\label{sec:rw_memory}

A growing literature studies long-term memory and historical evidence use in agents, with multimodal evidence only recently entering the picture, building on broader advances in multimodal model architectures and efficiency~\cite{long2025revisiting,zhao2026seeingendstepzero}, and with system-level reliability and security concerns emerging in adjacent agent domains~\cite{zhang2025enhancing,wang2026magesafeguardingllmagents,lin2026safeharness}. Architectures including Generative Agents~\cite{park2023generativeagents}, Reflexion~\cite{shinn2023reflexion}, MemGPT~\cite{packer2023memgpt}, and CogMem~\cite{zhang2025cogmem} use retrieval, reflection, virtual context, or layered persistent memory for long-horizon reasoning.
More recent multimodal benchmarks extend this direction: M3-Agent~\cite{long2026m3agent} studies entity-level memory from continuous video in a single-session robotic/video setting, alongside progress on multimodal introspective reasoning over long-form video~\cite{li2026revisor}; ATM-Bench~\cite{mei2026atmbench}
introduces personalized multimodal memory QA over long user histories but does not control artifact access scenario; and Mem-Gallery~\cite{memgallery2026} combines multimodal content with multi-session dialogue
evaluation. By contrast, major memory benchmarks including LoCoMo~\cite{maharana2024locomo}, LongMemEval~\cite{wu2025longmemeval}, MemBench~\cite{membench2025}, and
MemoryArena~\cite{memoryarena2026} remain primarily text-only. MIRAGE focuses on a narrower problem: whether an agent can re-access and use historical evidence
objects under conversation-state variation; Appendix~\ref{app:benchmark_depth} provides detailed benchmark and depth-normalized comparisons.

\subsection{Retrieval Faithfulness and Memory Hallucination}\label{sec:rw_faithful}

A separate line of work examines whether agents that invoke retrieval tools actually use the retrieved content, and when they decide to invoke tools at all~\cite{liu2026llms}, including adaptive multimodal search~\cite{wu2026promsa} and multi-agent QA for open-domain multi-hop reasoning~\cite{liu2026prisma}.
Despite RAG becoming a standard grounding approach~\cite{gao2023ragsurvey}, models can be distracted by irrelevant retrieved context~\cite{shi2023irrelevant}, and Self-RAG~\cite{asai2024selfrag} shows that self-reflection on retrieval relevance helps but remains single-turn and text-only.
Ma et al.~\cite{ma2025proofofuse} identify \emph{tool-call hacking} in multi-source RAG agents, the closest precedent to our notion of hallucinated grounding, though targeting general pipelines rather than multimodal personal agents.
HaluMem~\cite{halumem2025} benchmarks hallucination in agent memory operations but covers text-only settings without stratifying by access scenario, and broader surveys~\cite{huang2025hallucination_survey,hallucination_survey2025} propose taxonomies without operationalizing them under conversation-state variation.
We extend these findings to the cross-session multimodal setting.

\subsection{Access Under Long Context}\label{sec:rw_longctx}

Even when the target artifact remains within the active context, the agent may fail to use it.
Needle-in-a-Haystack~\cite{kamradt2023needle} and Lost in the Middle~\cite{liu2024lost} show systematic mid-context retrieval degradation; RULER~\cite{hsieh2024ruler} and $\infty$Bench~\cite{zhang2024infinitybench} confirm that effective context size is often far smaller than the advertised window.
In agentic settings, web-agent evaluations~\cite{webagent2025longcontext} report success rates collapsing from 40--50\% to under 10\% in long-context conditions.
Levy et al.~\cite{contextlength2025emnlp} further show that extending context damages reasoning even when the model can still locate evidence.
These findings motivate our S1 baseline: in-window availability does not guarantee in-context utilization.

%%% ============================================================
\section{Methodology}\label{sec:method}
%%% ============================================================

The agent system is organized as a recurring interaction loop around a multimodal backbone: it receives user messages and multimodal observations, prepares active context with memory and tool affordances, invokes the backbone, and executes any tool requests before returning the response.
Continuity depends on the backbone and how the surrounding system maintains state, exposes memory, and mediates retrieval across time.

We distinguish two memory surfaces: the active context $c_t$ (model-visible same-session content at turn $t$) and the persistent workspace $\mathcal{W}$ (durable files, notes, and extracted records not continuously visible).
When a query depends on past evidence, the agent may reach it through \textbf{$\Rcontext$} (same-session context continuity) or \textbf{$\Rtool$} (explicit tool-mediated retrieval).\label{sec:agent_arch}

\begin{figure*}[t]
  \centering
  \includegraphics[width=\textwidth]{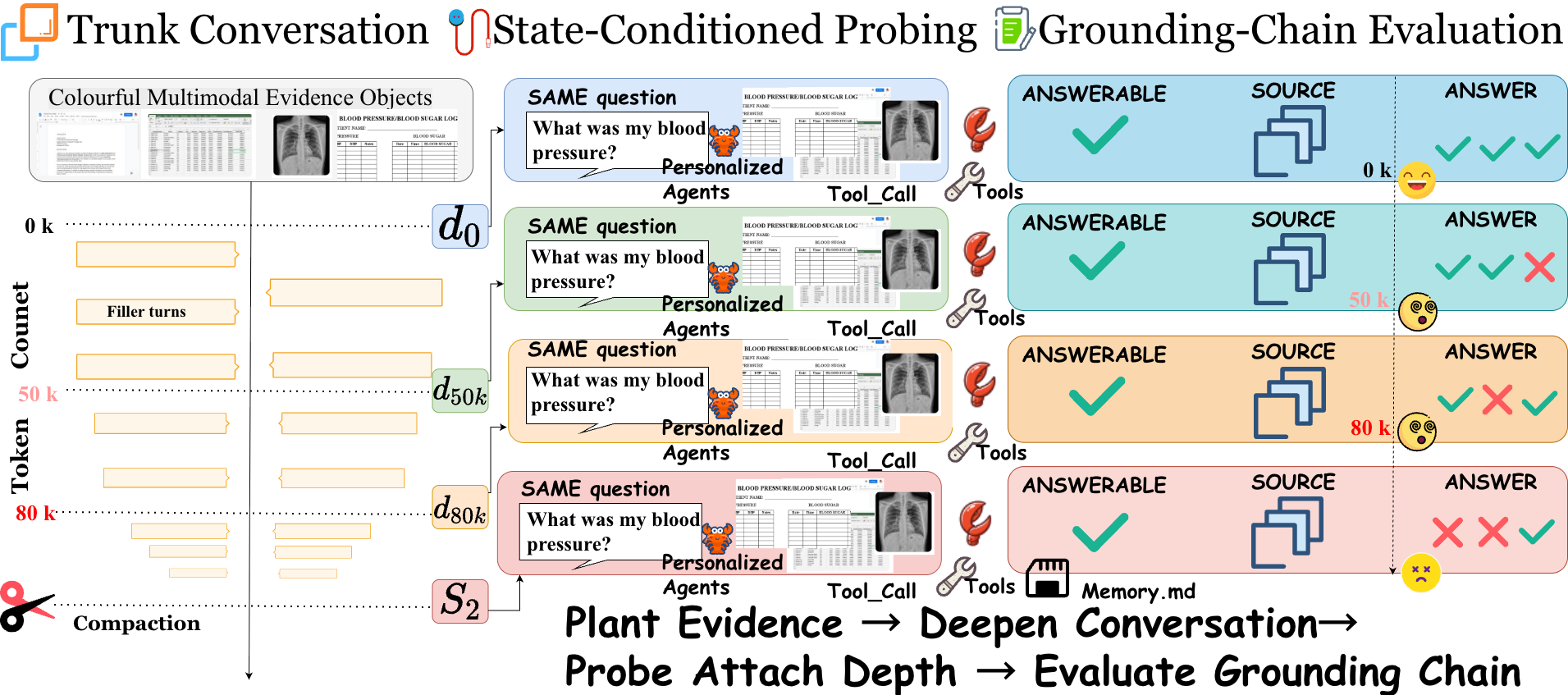}
  \caption{MIRAGE evaluation pipeline. \textbf{Left}: a trunk conversation plants multimodal evidence objects, then filler turns deepen context through four checkpoints (S1-d0, S1-d50k, S1-d80k, S2). \textbf{Center}: the same probe question is issued at each state to a personal agent with tool access. \textbf{Right}: responses are evaluated on a three-stage grounding chain (answerability, source attribution, answer correctness), producing state-conditioned diagnostic profiles.}
  \label{fig:overview}
\end{figure*}

\subsection{Problem Statement}\label{sec:problem}

We study historical evidence use under conversation-state variation: given a reference query $q$ about a prior evidence object $a^\star$, the agent must determine whether the question is answerable from its history, recover the correct supporting evidence if it exists, and answer from that evidence rather than from a plausible guess.
Let the persistent memory surface be $\mathcal{W} = \{a_1, a_2, \dots, a_K\}$ and let the agent act under policy
\begin{equation}
  \pi : (c_t,\; q,\; \mathcal{W}) \;\mapsto\; (\mathbf{u},\; y),
\end{equation}
where $\mathbf{u}$ is the action/tool trace and $y$ is the final response.
The core question is whether $y$ is \emph{authentically grounded} in $a^\star$ rather than merely plausible.
We define grounding faithfulness using correct source identification and a correct normalized value:
\begin{equation}\label{eq:grounded}
  \text{Grounded}(y;\,a^\star,q) {=} 1 \;\iff\; \\
  (\mathrm{SC} {=} 1) \,\wedge\, (\mathrm{VC} {=} 1).
\end{equation}
$\mathrm{SC}$ denotes source correctness, and $\mathrm{VC}$ denotes value correctness after normalization.

\subsection{Historical Artifact Planting}\label{sec:planting}

Figure~\ref{fig:overview} summarizes the pipeline: a trunk conversation plants multimodal evidence, filler turns deepen context through four checkpoints, and the same probe questions are evaluated at each state on a three-stage grounding chain.
The central experimental constraint is that evidence objects and questions are held fixed; only the \emph{state} changes.
This is a deliberate design choice: if both the evidence set and the conversation state varied simultaneously, observed performance differences could reflect evidence-level difficulty rather than state-conditioned access failure.
By fixing a small, controlled evidence set and reusing it across all states and models, MIRAGE isolates conversation state as the sole experimental variable, prioritizing internal validity and causal interpretability over evidence-set breadth.

\noindent\textbf{Evidence objects, conversations, and filler turns.}
The benchmark begins by planting a fixed set of evidence objects $\mathcal{A} = \{a_1, \dots, a_K\}$ into the agent's interaction history, each carrying a stable identifier and modality-aware content.
These objects are introduced during an initial conversation $C$ consisting of alternating user--agent turns.
After evidence planting, MIRAGE appends filler turns to increase input depth and induce native compaction.

\subsection{State Construction}\label{sec:setting}

\noindent\textbf{Trunk conversation and checkpoints.}
The shared trunk conversation is the composition
\begin{equation}
  T = C_{\mathrm{plant}} \oplus F,
\end{equation}
where $C_{\mathrm{plant}}$ contains the evidence-introduction turns and $F$ is the subsequent filler suffix.
At selected prefixes of $T$, the runner materializes checkpoints
\begin{equation}
  \mathcal{K} = \{\kappa_1, \kappa_2, \dots, \kappa_J\},
\end{equation}
where each $\kappa_j$ is captured from $T_{\le \tau_j}$ and stores the runtime state needed for later resumption.

\noindent\textbf{Depth, states, and probes.}
We define checkpoint depth by the effective-input-token function
\begin{equation}
  d(\kappa_j) = \mathrm{EIT}(\kappa_j),
\end{equation}
where $\mathrm{EIT}(\kappa_j)$ is the provider-reported effective input tokens at capture time.
Increasing depth is induced by extending the filler suffix $F$.
A \textbf{state} is the evaluation condition associated with a resumed execution from a checkpoint.
A state map $\psi : \mathcal{S} \to \mathcal{K}$ assigns each named state to one checkpoint.
The current study uses four named states:
\begin{equation}
  \mathcal{S} = \{\texttt{S1-d0},\; \texttt{S1-d50k},\; \texttt{S1-d80k},\; \texttt{S2}\}.
\end{equation}
The first three form the pre-compaction same-session regime, \texttt{S1} $= \{\texttt{S1-d0}, \texttt{S1-d50k}, \texttt{S1-d80k}\}$, while \texttt{S2} is the post-compaction same-session state.
These four points sample three qualitatively distinct regions of the context lifecycle: near the planted evidence (\texttt{S1-d0}), at mid-range and near the context-window boundary (\texttt{S1-d50k}, \texttt{S1-d80k}), and after the first native compaction event (\texttt{S2}).
A finer-grained depth sweep would increase cost linearly but is unlikely to reveal regime transitions beyond the pre-/post-compaction boundary that the current design already captures.
These coordinates are runtime-grounded proxies, not backbone-normalized difficulty bins: models with different context-use strategies, tool policies, and compaction behaviors may experience the same EIT milestone differently.
Multiple models are evaluated to test whether state-conditioned patterns generalize across architecturally diverse backbones; shared state coordinates enable both within-model state comparisons and cross-model pattern validation under the same runtime conditions.

A probe is the tuple
\begin{equation}
  p_{i,s} = (\psi(s),\; q_i), \qquad s \in \mathcal{S},
\end{equation}
where $q_i$ is one base question and $\psi(s)$ is the checkpoint associated with state $s$.
If the bank contains $N$ base questions, the full evaluation contains $N \times |\mathcal{S}|$ probes.
Algorithm~\ref{alg:mirage_pipeline} summarizes the full pipeline.

\begin{algorithm}[t]
\caption{MIRAGE Conversation-State Evaluation Pipeline}
\label{alg:mirage_pipeline}
\begin{algorithmic}[1]
\REQUIRE planted evidence set $\mathcal{A}$, planting conversation $C_{\mathrm{plant}}$, filler turns $F$, state set $\mathcal{S}$, question bank $\mathcal{Q}$
\STATE Build trunk conversation $T \leftarrow C_{\mathrm{plant}} \oplus F$
\STATE Materialize checkpoints $\mathcal{K}$ from $T$ at selected depth or post-compaction boundaries
\FOR{each state $s \in \mathcal{S}$}
  \FOR{each base question $q_i \in \mathcal{Q}$}
    \STATE Fresh-restore checkpoint $\kappa \leftarrow \psi(s)$ \COMMENT{independent per probe}
    \STATE Issue probe $p_{i,s} = (\psi(s), q_i)$
    \STATE Record structured response $\hat{g}_{i,s}$ and runtime trace
    \STATE Compute deterministic outcome metrics and retrieval-path diagnostics
  \ENDFOR
\ENDFOR
\RETURN conversation-state score table and diagnostic traces
\end{algorithmic}
\end{algorithm}

\subsection{Unified Probe Protocol}\label{sec:threestage}

Critically, each probe is issued on a freshly restored copy of the checkpoint, so that no probe observes the question or answer of any other probe.
This guarantees per-probe independence and eliminates cross-probe contamination from accumulated conversation history.
Rather than splitting the study into separate task families, MIRAGE uses a single structured-response protocol that captures the full grounding chain in one turn.
Each probe requires the model to return:
\begin{quote}
\small
\texttt{ANSWERABLE=YES | NO}\\
\texttt{SOURCE=<artifact\_id or NONE>}\\
\texttt{ANSWER=<short value or NONE>}
\end{quote}
This design lets one response reveal whether the model judges the question answerable, whether it identifies the correct source, and whether it extracts the right content, without requiring separate probing stages.
Under \texttt{Cm}, the same protocol also exposes whether the model can express retrieval through the executable tool interface used by the current runtime.

\subsection{Grounding-Oriented Evaluation}\label{sec:evaluation}

\noindent\textbf{Deterministic Scoring.}\label{sec:metrics}
For each probe, the model returns a structured response
\begin{equation}
  \hat{g}_{i,s} = (\hat{z}_{i,s},\; \widehat{\mathrm{src}}_{i,s},\; \hat{y}_{i,s}),
\end{equation}
consisting of a predicted answerability judgment, source identifier, and answer value.
We compare these against the gold annotation and compute six per-probe indicators over the answerable subset $\mathcal{Q}^{+}$ or the unanswerable subset $\mathcal{Q}^{-}$ as appropriate.
The first-letter abbreviations below serve as column headers throughout the experiment tables:
\begin{align}
  \mathrm{AC}(i,s) &= \mathbf{1}[\hat{z}_{i,s}=z_i^\star], \label{eq:ac} \\
  \mathrm{SC}(i,s) &= \mathbf{1}[\mathrm{canon}(\widehat{\mathrm{src}}_{i,s})=\mathrm{src}_i^\star], \label{eq:sc} \\
  \mathrm{VC}(i,s) &= \mathbf{1}[\mathrm{norm}(\hat{y}_{i,s})=y_i^\star], \label{eq:vc} \\
  \mathrm{GC}(i,s) &= \mathrm{SC}(i,s) \cdot \mathrm{VC}(i,s), \label{eq:gc} \\
  \mathrm{HR}(i,s) &= \mathbf{1}[\hat{z}_{i,s}{=}\texttt{YES} \;\lor\; \hat{y}_{i,s}{\neq}\texttt{NONE}], \label{eq:hr} \\
  \mathrm{WS}(i,s) &= \mathbf{1}[\mathrm{canon}(\widehat{\mathrm{src}}_{i,s}) \neq \mathrm{src}_i^\star], \label{eq:ws}
\end{align}
\textbf{AC}, \textbf{GC}, \textbf{HR}, and \textbf{WS} denote answerability correctness, grounded correctness, hallucination rate, and wrong-or-missing source rate, respectively.
AC is computed over all probes; SC, VC, GC, and WS over the answerable subset $\mathcal{Q}^{+}$; and HR over the unanswerable subset $\mathcal{Q}^{-}$.
Consistent with the released scorer, WS counts every canonicalized source mismatch, including \texttt{NONE}; WS is the complement of SC on the same scored subset.
We report state-level averages by averaging each indicator over its relevant subset per state.

All matching is deterministic.
Enum fields are matched after case normalization; source identifiers are canonicalized to collapse artifact paths, derived files, screenshot summaries, and dated memory notes onto the underlying evidence identifier; numeric and short string answers are normalized with fixed rules.
\textbf{PF} denotes parse failure and is tracked separately for outputs that do not satisfy the full structured protocol. Under \texttt{Cm}, pseudo-tool outputs and malformed tool syntax are therefore part of the measured behavior rather than scorer noise.

A \emph{mirage} is an answerable probe for which the model predicts \texttt{YES} without achieving grounded correctness. Its state-level rate is
\begin{equation}
  \mathrm{MI}_{s}=
  \frac{1}{|\mathcal{Q}^{+}_{s,\mathrm{obs}}|}
  \sum_{i\in\mathcal{Q}^{+}_{s,\mathrm{obs}}}
  \mathbf{1}[\hat{z}_{i,s}=\texttt{YES}\ \land\ \mathrm{GC}(i,s)=0],
\end{equation}
where $\mathcal{Q}^{+}_{s,\mathrm{obs}}$ contains all recorded answerable calls and parse failures contribute zero. MI comprises outcome-preserving claimed-answerability failures ($\hat{z}{=}\texttt{YES}$, VC${=}1$, GC${=}0$) and answer-failure mirages ($\hat{z}{=}\texttt{YES}$, VC${=}0$). We separately report $\mathrm{OG}_{s}=\mathrm{VC}_{s}-\mathrm{GC}_{s}$ over valid parsed answerable probes; it measures correct-but-ungrounded outputs irrespective of $\hat{z}$.

\noindent\textbf{Retrieval-Path Diagnostics.}\label{sec:threeaxis}
Beyond outcome scores, MIRAGE records the observed retrieval path for every recorded probe call: whether the agent accesses evidence through same-session context continuity ($\Rcontext$) or through an explicit tool call ($\Rtool$), following the definitions in Section~\ref{sec:agent_arch}.
These labels serve a mechanistic role rather than a headline one: they reveal whether state change alters \emph{how} evidence is accessed, even when top-line accuracy does not shift dramatically.

%%% ============================================================
\section{Experiments}\label{sec:experiments}
%%% ============================================================

\subsection{Setup}\label{sec:setup}

\noindent\textbf{Datasets.}
The corpus contains six planted multimodal evidence objects drawn from ChartQA~\cite{masry-etal-2022-chartqa} and ScreenSpot~\cite{cheng2024seeclick}, together with their derived workspace representations.
The 200-question bank pairs 100 answerable items with 100 domain- and format-matched unanswerable counterparts; gold labels record answerability, canonical source, and normalized answer, while artifact metadata records difficulty, query type, and shortcut risk (Appendix~\ref{app:data_controls}).
Across four states, this yields 800 scored probes per model--condition pair.
\revadd{Q1}{Appendix~\ref{app:evidence_extension} reports a 109-probe extension across six additional evidence types.}
\revadd{Q8}{Each cell is averaged over three independent runs; Appendix~\ref{app:uncertainty} reports representative confidence intervals and routing counts.}

\noindent\textbf{Models.}
We evaluate two frontier API references under \texttt{C0}, GPT-5~\cite{openai2025gpt5} and Claude-4.5-Haiku~\cite{anthropic2025claudehaiku}, and five open-weight VLMs: Qwen3-VL-4B/8B/30B~\cite{bai2025qwen3vl}, InternVL3.5-20B (MoE)~\cite{wang2025internvl35}, and Gemma-3-27B~\cite{gemma2025gemma3}.
\texttt{C0} permits default evidence-access behavior, whereas \texttt{Cm} requires \texttt{artifact\_recall} before answering and is applied to the locally deployable open-model group whose \texttt{C0} degradation motivates the mitigation study.
\texttt{Cm} is a system-level condition under the deployed runtime contract; oracle or forced execution would test a different, component-level question.

\noindent\textbf{Metrics and diagnostics.}
We report the deterministic indicators defined in Section~\ref{sec:metrics}: \textbf{AC}, \textbf{SC}, \textbf{VC}, \textbf{GC}, \textbf{HR}, \textbf{WS}, and \textbf{PF}, and record retrieval paths ($\Rcontext$ vs $\Rtool$) as mechanistic labels.
For state analysis, \textbf{DS}${=}100(\mathrm{GC}_{d0}-\mathrm{GC}_{d80k})/\mathrm{GC}_{d0}$ is the relative GC reduction from \texttt{S1-d0} to \texttt{S1-d80k}; \revadd{Q8}{$\Dcomp{=}\mathrm{GC}_{S2}-\mathrm{GC}_{d80k}$ is the net GC change}; and \textbf{Ret} is the fraction of \texttt{S1-d0}-correct probes remaining correct at \texttt{S1-d80k}.
MI and OG follow Section~\ref{sec:metrics}. Table~\ref{tab:diagnostics} reports $\mathrm{MI}_{d80k}$ in percent (the \texttt{S1-d80k} values in Figure~\ref{fig:mi_heatmap} multiplied by 100).

\noindent\textbf{Implementation details.}
All models share the same OpenClaw-based~\cite{openclaw2026} runtime, checkpoint family, and probe pipeline.
We use effective input tokens (EIT) reported by backbone providers as the depth signal and native \texttt{compactionCount} to detect compaction events, instantiating the four abstract states as \texttt{S1-d0} (EIT 23{,}184, 0 compactions), \texttt{S1-d50k} (50{,}217, 0), \texttt{S1-d80k} (80{,}436, 0), and \texttt{S2} (100{,}081, 1 compaction).
\revadd{Q3}{Fixed-seed, topic-disjoint filler turns only increase input depth and trigger native compaction, whose runtime policy remains unmodified; Appendix~\ref{app:data_controls} details the semantic controls.}
\revadd{Q2}{Appendix~\ref{app:compaction_ablation} reports the controlled \texttt{S2} ablation over GPT-5 summary formats.}

\noindent\textbf{Scope and generalizability.}
MIRAGE evaluates context depth, native compaction, and tool-mediated retrieval in a controlled planted-evidence setting; its results characterize state-conditioned behavior rather than provide a coverage-complete estimate of multimodal personal-agent workloads.
Because the state coordinates are runtime-defined rather than backbone-normalized, the post-compaction results apply to the model--runtime stack and do not isolate the contributions of memory-surface design, tool orchestration, provider EIT definitions, or compaction summary quality (Appendix~\ref{app:benchmark_depth}).

%%% ============================================================
\subsection{Main Results}\label{sec:results_main}
%%% ============================================================

\begin{table}[t]
  \centering
  \scriptsize
  \setlength{\tabcolsep}{2pt}
  \renewcommand{\arraystretch}{0.88}
  \caption{Baseline (\texttt{C0}) metrics across all seven backbones and four conversation states.
    Metric columns are percentages; PF is a count.
    The gap VC$-$GC reveals overestimation by outcome-only evaluation.}
  \label{tab:c0_main}
  \resizebox{\columnwidth}{!}{%
  \begin{tabular}{ll rrrrrr r}
    \toprule
    Model & State & AC & SC & VC & GC & HR & WS & PF \\
    \midrule
    \rowcolor{gray!15} \multicolumn{9}{l}{\textit{Frontier API (C0 only)}} \\
    \midrule
    \multirow{4}{*}{GPT-5~\cite{openai2025gpt5}}
      & \texttt{S1-d0}   & 85.1 & 93.9 & 85.7 & \textbf{85.7} & 22.7 &  6.1 &  5 \\
      & \texttt{S1-d50k} & 84.2 & 93.9 & 83.7 & 83.7 & 24.5 &  6.1 &  4 \\
      & \texttt{S1-d80k} & 83.9 & 93.9 & 83.8 & 82.8 & 25.0 &  6.1 &  1 \\
      & \texttt{S2}      & 88.4 & 97.0 & 85.9 & 84.8 & 19.2 &  3.0 &  2 \\
    \midrule
    \multirow{4}{*}{Claude-4.5-Haiku~\cite{anthropic2025claudehaiku}}
      & \texttt{S1-d0}   & 80.7 & 94.9 & 46.5 & \textbf{46.5} & 30.6 &  5.1 &  3 \\
      & \texttt{S1-d50k} & 73.2 & 83.7 & 43.9 & 43.9 & 42.0 & 16.3 &  2 \\
      & \texttt{S1-d80k} & 71.4 & 75.0 & 42.0 & 41.0 & 39.4 & 25.0 &  1 \\
      & \texttt{S2}      & 73.0 & 73.0 & 40.0 & 36.0 & 39.0 & 27.0 &  0 \\
    \midrule
    \rowcolor{gray!15} \multicolumn{9}{l}{\textit{Open VLM (C0 only)}} \\
    \midrule
    \multirow{4}{*}{Qwen3-VL-4B~\cite{bai2025qwen3vl}}
      & \texttt{S1-d0}   & 77.4 & 71.0 & 61.0 & \textbf{54.0} & 27.3 & 29.0 &  1 \\
      & \texttt{S1-d50k} & 61.0 & 31.6 & 31.6 & 18.4 & 52.6 & 68.4 &  3 \\
      & \texttt{S1-d80k} & 55.0 &  3.0 &  5.0 &  1.0 & 72.0 & 97.0 &  0 \\
      & \texttt{S2}      & 64.9 & 52.1 & 44.8 & 34.4 & 53.1 & 47.9 &  5 \\
    \midrule
    \multirow{4}{*}{Qwen3-VL-8B~\cite{bai2025qwen3vl}}
      & \texttt{S1-d0}   & 70.5 & 89.5 & 72.1 & \textbf{69.8} & 57.5 & 10.5 &  0 \\
      & \texttt{S1-d50k} & 54.5 & 78.0 & 60.0 & 54.0 & 89.0 & 22.0 &  0 \\
      & \texttt{S1-d80k} & 57.0 & 27.0 & 47.0 & 18.0 & 82.0 & 73.0 &  0 \\
      & \texttt{S2}      & 57.5 & 79.0 & 46.0 & 44.0 & 78.0 & 21.0 &  0 \\
    \midrule
    \multirow{4}{*}{Qwen3-VL-30B~\cite{bai2025qwen3vl}}
      & \texttt{S1-d0}   & 71.5 & 82.0 & 74.0 & \textbf{72.0} & 42.0 & 18.0 &  0 \\
      & \texttt{S1-d50k} & 70.5 & 72.0 & 62.0 & 61.0 & 33.0 & 28.0 &  0 \\
      & \texttt{S1-d80k} & 65.5 & 47.0 & 44.0 & 41.0 & 20.0 & 53.0 &  0 \\
      & \texttt{S2}      & 63.5 & 53.1 & 39.8 & 37.8 & 28.3 & 46.9 &  3 \\
    \midrule
    \multirow{4}{*}{InternVL3.5-20B~\cite{wang2025internvl35}}
      & \texttt{S1-d0}   & 52.9 & 75.3 & 24.7 & \textbf{20.0} & 98.9 & 24.7 & 26 \\
      & \texttt{S1-d50k} & 52.3 &  1.1 &  3.4 &  1.1 & 80.0 & 98.9 & 28 \\
      & \texttt{S1-d80k} & 54.1 &  0.0 &  0.0 &  0.0 & 81.5 &100.0 & 78 \\
      & \texttt{S2}      & 56.6 & 60.6 & 18.1 & 13.8 & 88.6 & 39.4 & 18 \\
    \midrule
    \multirow{4}{*}{Gemma-3-27B~\cite{gemma2025gemma3}}
      & \texttt{S1-d0}   & 55.5 & 96.0 & 57.0 & \textbf{57.0} & 84.0 &  4.0 &  0 \\
      & \texttt{S1-d50k} & 56.0 & 71.0 & 45.0 & 34.0 & 86.0 & 29.0 &  0 \\
      & \texttt{S1-d80k} & 57.3 & 70.7 & 38.4 & 29.3 & 82.0 & 29.3 &  1 \\
      & \texttt{S2}      & 54.0 & 65.0 & 43.0 & 29.0 & 88.0 & 35.0 &  0 \\
    \bottomrule
  \end{tabular}}%
\end{table}

% Table~\ref{tab:c0_main} summarizes the \texttt{C0} baseline.
% The remainder of this section unpacks three recurring patterns: a large frontier--open separation in grounding fidelity, heterogeneous degradation among open models, and a non-monotonic relation between \texttt{S1-d80k} and \texttt{S2}.

%%% ============================================================
\subsubsection{RQ1: What Failure Patterns Emerge Under Conversation-State Variation?}\label{sec:rq1}
%%% ============================================================

\noindent\textbf{Frontier models are stable but internally divergent.}
Table~\ref{tab:diagnostics} shows that GPT-5 has minimal DS at 3.4\% and retains 94.0\% of its \texttt{S1-d0}-correct answers at \texttt{S1-d80k}; $\Dcomp{=}{+}0.02$ indicates slight \texttt{S2} recovery.
Claude-4.5-Haiku degrades more clearly: DS${=}$11.8\%, retention drops to 73.9\%, and $\Dcomp{=}{-}0.05$ shows continued \texttt{S2} decline.
In Table~\ref{tab:c0_main}, WS rises from 5.1 to 27.0, indicating that Claude-4.5-Haiku's failures increasingly manifest as provenance confusions rather than refusals.

\noindent\textbf{Open models show larger and more heterogeneous degradation.}
Table~\ref{tab:diagnostics} shows more heterogeneous degradation in the open-model group: Qwen3-VL-30B (43.1\%) $<$ Gemma-3-27B (48.6\%) $<$ Qwen3-VL-8B (74.2\%) $<$ Qwen3-VL-4B (98.1\%) $<$ InternVL3.5-20B (100\%).
The most extreme retention failures occur at the low end: Qwen3-VL-4B retains 1 of 54 \texttt{S1-d0}-correct answers at \texttt{S1-d80k} (Ret${=}$1.9\%), and InternVL3.5-20B retains none (Ret${=}$0\%).

\begin{figure}[t]
  \centering
  \includegraphics[width=\columnwidth]{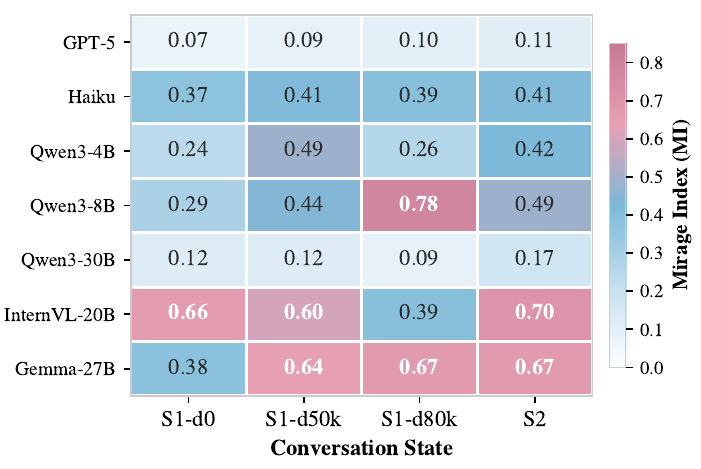}
  \caption{Mirage Index (MI; Section~\ref{sec:metrics}) across models and states. Higher values indicate a larger fraction of recorded answerable calls for which the model claims answerability without achieving grounded correctness.}
  \label{fig:mi_heatmap}
\end{figure}

\noindent\textbf{The mirage phenomenon.}
Figure~\ref{fig:mi_heatmap} highlights states in which models claim answerability without achieving grounded correctness.
The strongest case is Qwen3-VL-8B at \texttt{S1-d80k}: 78\% of recorded answerable calls jointly satisfy $\hat{z}{=}\texttt{YES}$ and GC${=}0$ (MI${=}$0.78); the marginal YES and GC rates are 96\% and 18\%, respectively.
As defined in Section~\ref{sec:metrics}, MI covers outcome-preserving and answer-failure mirages. OG is reported separately because it ignores the answerability prediction: here OG${=}$0.29 means that answer correctness overestimates grounded correctness by 29 percentage points.
InternVL3.5-20B at \texttt{S2} (MI${=}$0.70) and Gemma-3-27B at \texttt{S1-d80k}/\texttt{S2} (MI${=}$0.67) exhibit high mirage rates. By contrast, GPT-5 maintains MI${\le}$0.11 and OG${\le}$0.01 across all states.

\begin{figure*}[t]
  \centering
  \includegraphics[width=\textwidth]{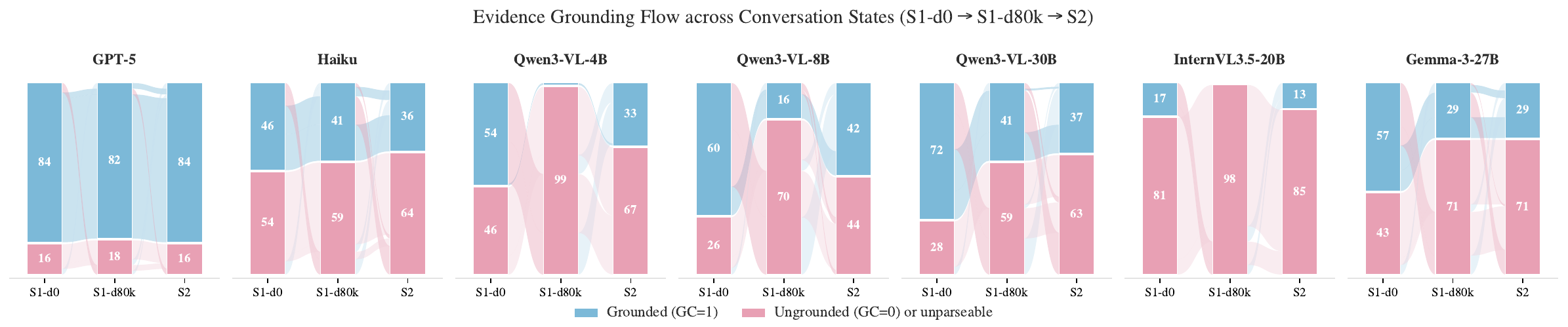}
  \caption{Evidence-grounding transitions across states for all seven models. Numbers show counts over answerable probes shared across the three states; blue denotes grounded outputs, and pink denotes ungrounded or unparseable outputs.}
  \label{fig:alluvial}
\end{figure*}

\noindent\textbf{S2 is not simply a harder version of S1-d80k.}
Figure~\ref{fig:alluvial} shows that post-compaction behavior is qualitatively distinct from deep pre-compaction failure.
GPT-5 shows near-complete stability (76/100 $\checkmark{\to}\checkmark{\to}\checkmark$).
Qwen3-VL-8B exhibits a prominent recovery flow: 26 questions that fail at \texttt{S1-d80k} recover at \texttt{S2} ($\checkmark{\to}\times{\to}\checkmark$), consistent with the post-compaction representation in the current runtime re-surfacing some evidence.
InternVL3.5-20B shows limited recovery, with only 13 of 98 probes grounded at \texttt{S2}.
$\Dcomp$ quantifies this divergence: Qwen3-VL-4B recovers substantially at ${+}0.33$, while Claude-4.5-Haiku continues to degrade at ${-}0.05$.
This separates \texttt{S1-d80k} and \texttt{S2} as distinct failure regimes.

\noindent\textbf{Degradation is model- and state-dependent.}
Table~\ref{tab:c0_main} shows that Qwen3-VL-4B and 8B deteriorate sharply by \texttt{S1-d80k} but partially recover at \texttt{S2}, whereas Qwen3-VL-30B degrades more gradually and continues to decline after compaction.
These contrasting trajectories reinforce that deep pre-compaction and post-compaction states constitute distinct failure regimes rather than points on a single monotonic degradation curve. The extension set shows the same deep-state reduction in source correctness across all three Qwen sizes (Appendix~\ref{app:evidence_extension}).

\begin{table}[t]
  \centering
  \scriptsize
  \setlength{\tabcolsep}{2pt}
  \renewcommand{\arraystretch}{0.88}
  \caption{Cross-state diagnostic indicators under \texttt{C0}.
    DS, Ret, and $\mathrm{MI}_{d80k}$ are percentages; $\Dcomp$ and OG are proportions.}
  \label{tab:diagnostics}
  \resizebox{\columnwidth}{!}{%
  \begin{tabular}{l rrrr rrrr}
    \toprule
    & & & & & \multicolumn{4}{c}{OG (per state)} \\
    \cmidrule(lr){6-9}
    Model & DS & $\Dcomp$ & Ret & $\mathrm{MI}_{d80k}$ & \texttt{S1-d0} & \texttt{S1-d50k} & \texttt{S1-d80k} & \texttt{S2} \\
    \midrule
    GPT-5~\cite{openai2025gpt5}           &  3.4 & $+$0.02 & 94.0 & 10.0 & 0.00 & 0.00 & 0.01 & 0.01 \\
    Claude-4.5-Haiku~\cite{anthropic2025claudehaiku}           & 11.8 & $-$0.05 & 73.9 & 39.0 & 0.00 & 0.00 & 0.01 & 0.04 \\
    Qwen3-VL-4B~\cite{bai2025qwen3vl}    & 98.1 & $+$0.33 &  1.9 & 26.0 & 0.07 & 0.13 & 0.04 & 0.10 \\
    Qwen3-VL-8B~\cite{bai2025qwen3vl}    & 74.2 & $+$0.26 & 21.7 & 78.0 & 0.02 & 0.06 & 0.29 & 0.02 \\
    Qwen3-VL-30B~\cite{bai2025qwen3vl}   & 43.1 & $-$0.03 & 51.4 &  9.0 & 0.02 & 0.01 & 0.03 & 0.02 \\
    InternVL3.5-20B~\cite{wang2025internvl35} &100.0 & $+$0.14 &  0.0 & 39.0 & 0.04 & 0.02 & 0.00 & 0.04 \\
    Gemma-3-27B~\cite{gemma2025gemma3}     & 48.6 & $-$0.003 & 42.1 & 67.0 & 0.00 & 0.11 & 0.09 & 0.14 \\
    \bottomrule
  \end{tabular}}%
\end{table}

%%% ============================================================
\subsubsection{RQ2: Why Do These Patterns Arise?}\label{sec:rq2}
%%% ============================================================

\begin{figure*}[t]
  \centering
  \includegraphics[width=\textwidth]{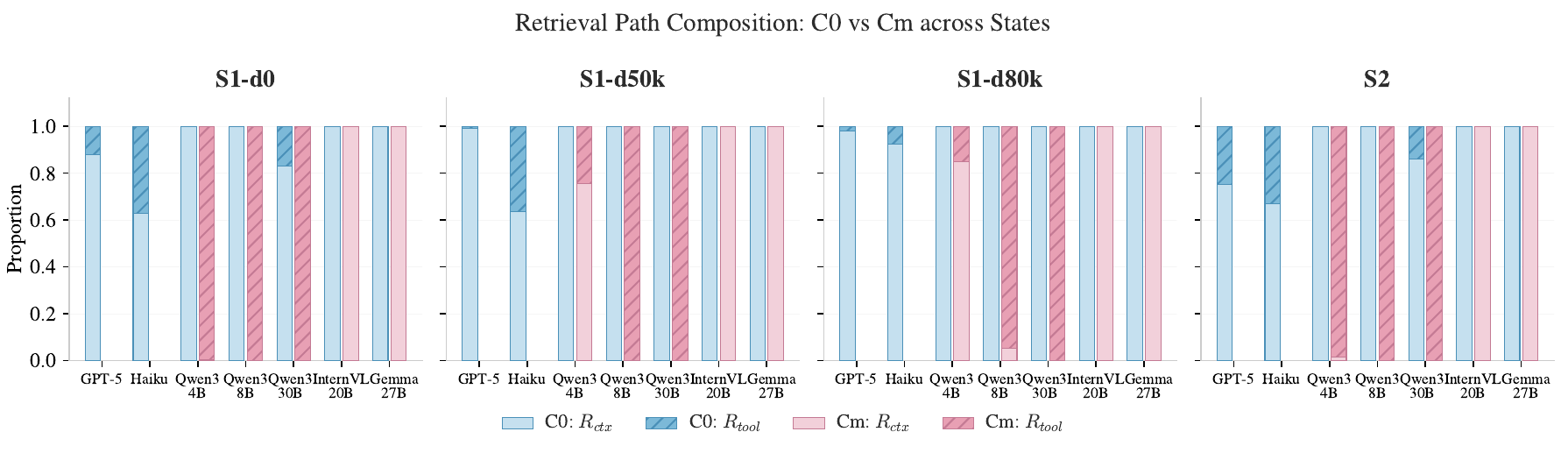}
  \caption{Retrieval-path composition under \texttt{C0} and \texttt{Cm}. Percentages use all recorded probe calls in each model--condition--state cell; blue/pink encode \texttt{C0}/\texttt{Cm}, and solid/hatched encode $\Rcontext$/$\Rtool$. Frontier models have \texttt{C0} only.}
  \label{fig:rpath}
\end{figure*}

\noindent\textbf{Frontier vs.\ open: a fundamental retrieval-mechanism divide.}
Figure~\ref{fig:rpath} reveals a stark asymmetry under \texttt{C0} in the current runtime.
The frontier models actively employ tool-mediated retrieval: GPT-5 routes 25\% of recorded \texttt{S2} calls through $\Rtool$, and Claude-4.5-Haiku uses tools for 33--37\% of recorded calls even at shallow depth.
Open models are predominantly $\Rcontext$-driven; Qwen3-VL-30B is the only partial exception, with $\Rtool$ at 17\% in \texttt{S1-d0} and 14\% in \texttt{S2}.
Even when \texttt{S1-d80k} WS reaches 97.0\% for Qwen3-VL-4B and 100.0\% for InternVL3.5-20B, neither shifts into $\Rtool$, suggesting that models in this setting do not recognize failed in-context access and switch retrieval routes.
Under \texttt{Cm}, Qwen3-VL-8B and 30B reach $\Rtool{=}95$--$100\%$ across states; 4B does so only at \texttt{S1-d0} and \texttt{S2}. InternVL3.5-20B and Gemma-3-27B remain at $\Rtool{=}0\%$, indicating noncompliance under the current orchestration contract.

\noindent\textbf{Source drift: even correct answers change evidence paths.}
Among probes where GC${=}$1 at both \texttt{S1-d0} and \texttt{S1-d80k}, 30--59\% use a different \texttt{pred\_source} across the two states (Qwen3-VL-30B: 59\%, Gemma-3-27B: 58\%, Qwen3-VL-8B: 54\%, Claude-4.5-Haiku: 44\%, GPT-5: 30\%).
State change therefore alters not only \emph{whether} the agent answers correctly, but also \emph{how} it reaches the correct evidence, a shift invisible to outcome-only evaluation.
This means that even when GC remains stable, the underlying evidence-access mechanism has silently shifted, an instability that could compound across longer interaction histories.

\begin{revblock}{Q2}
\noindent\textbf{Compaction ablation.}
Per fact citations raise Qwen3-VL-8B SC from 79 to 95 while GC remains near 45, whereas 4B and 30B gain no GC; summary format can improve attribution without reliably improving grounding (Appendix~\ref{app:compaction_ablation}).
\end{revblock}

%%% ============================================================
\subsubsection{RQ3: Can Retrieval Pressure Mitigate State-Dependent Provenance Failure?}\label{sec:rq3}
%%% ============================================================

\begin{figure*}[t]
  \centering
  \includegraphics[width=\textwidth]{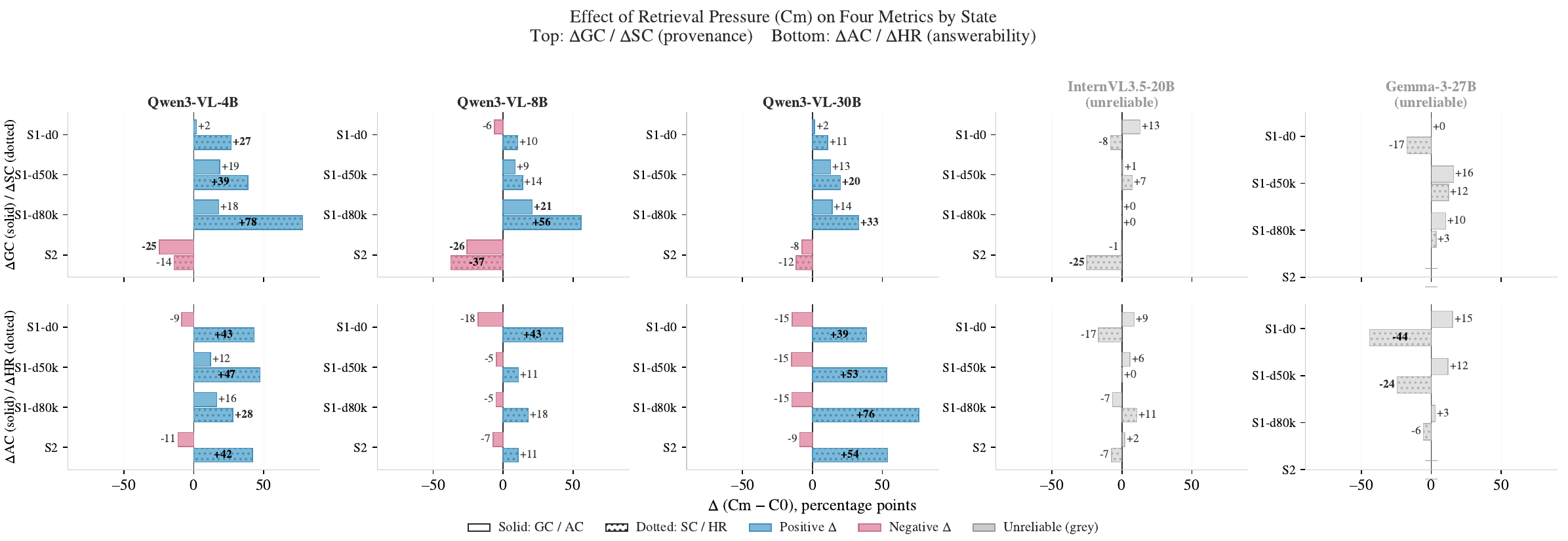}
  \caption{Effect of retrieval pressure, shown as $\Delta(\texttt{Cm}-\texttt{C0})$ in percentage points. Top: GC/SC; bottom: AC/HR. Grey marks tool non-compliance or parse failure; ``---'' denotes unevaluable cells.}
  \label{fig:cm_butterfly}
\end{figure*}

\noindent\textbf{Stage 1: Tool-compliance triage.}
Under \texttt{Cm}, all three Qwen models execute tool calls, although 4B remains predominantly $\Rcontext$-driven at \texttt{S1-d50k} and \texttt{S1-d80k}. The other models fail to produce executable calls: InternVL3.5-20B emits pseudo-tool text (e.g., \texttt{<|im\_start|>tool\,...\,artifact\_recall}), leaving $\Rtool{=}0\%$ under both conditions, while Gemma-3-27B's PF rises from 0.1\% to ${\ge}87\%$ and makes \texttt{S2} unevaluable. We treat both as mitigation-execution failures under the current orchestration contract, exclude their \texttt{Cm} deltas, and report a controlled parser-adapter check in Appendix~\ref{app:tool_compliance}. \textbf{Stage 2: Grounding efficacy among compliant models.}
Restricting to the three compliant Qwen models, Figure~\ref{fig:cm_butterfly} shows that \texttt{Cm} substantially improves pre-compaction provenance.
At \texttt{S1-d80k}, Qwen3-VL-30B gains $\Delta$SC${=}{+}33$ and $\Delta$GC${=}{+}14$.
Qwen3-VL-8B gains $\Delta$SC${=}{+}56$ and $\Delta$GC${=}{+}21$, while Qwen3-VL-4B gains $\Delta$SC${=}{+}78$ and $\Delta$GC${=}{+}18$.
These improvements confirm that, among models that successfully execute retrieval, explicit tool use reduces provenance confusion when the original evidence still exists in the workspace.

\noindent\textbf{S2 is not reliably repaired by retrieval pressure.}
At \texttt{S2}, all three regress in SC ($-12$, $-37$, $-14$) and GC ($-8$, $-26$, $-24$). This contrast shows that retrieval pressure helps while evidence is intact but does not reliably repair the post-compaction state produced by the current runtime. We hypothesize that retrieved workspace content becomes lossier or differently organized after compaction, anchoring models to incomplete evidence.

\noindent\textbf{HR reveals a provenance--conservatism tradeoff.}
\texttt{Cm} increases HR even where it improves grounding: on Qwen3-VL-30B, $\Delta$HR reaches ${+}76$ at \texttt{S1-d80k} (Figure~\ref{fig:cm_butterfly}, bottom row).
The model becomes better at grounding answerable questions, but worse at refusing unsupported ones.

%%% ============================================================
\subsection{Case Study}\label{sec:casestudy}
%%% ============================================================

We illustrate two core findings with probe-level examples drawn from the evaluation traces (Table~\ref{tab:casestudy}).

\begin{table}[t]
  \centering
  \scriptsize
  \setlength{\tabcolsep}{2pt}
  \renewcommand{\arraystretch}{0.95}
  \caption{Two probe-level case studies: Case~1 is an outcome-preserving mirage; Case~2 shows that \texttt{Cm} repairs \texttt{S1-d80k} but not \texttt{S2}.}
  \label{tab:casestudy}
  \resizebox{\columnwidth}{!}{%
  \begin{tabular}{llllll r}
    \toprule
    Case & Cond. & State & Ans. & Source & Answer & GC \\
    \midrule
    \rowcolor{gray!15} \multicolumn{7}{l}{\textit{Case 1: Mirage (Qwen3-VL-8B, ``What is the username in the fitness profile?'')}} \\
    \midrule
    & C0 & \texttt{S1-d0}   & YES & \texttt{profile\ldots json} & leelin & \cellcolor{blue!12}\textbf{1} \\
    & C0 & \texttt{S1-d80k} & YES & \texttt{NONE}               & leelin & \cellcolor{red!12}\textbf{0} \\
    & C0 & \texttt{S2}      & YES & \texttt{profile\ldots json} & leelin & \cellcolor{blue!12}\textbf{1} \\
    \midrule
    \rowcolor{gray!15} \multicolumn{7}{l}{\textit{Case 2: Cm boundary (Qwen3-VL-30B, GDP query, gold: \texttt{cqa\_9f9df328})}} \\
    \midrule
    & C0 & \texttt{S1-d80k} & YES & \texttt{NONE}                   & NONE   & \cellcolor{red!12}\textbf{0} \\
    & Cm & \texttt{S1-d80k} & YES & \texttt{cqa\_9f9df328.png}      & 516.40 & \cellcolor{blue!12}\textbf{1} \\
    & Cm & \texttt{S2}      & YES & \texttt{NONE}                   & NONE   & \cellcolor{red!12}\textbf{0} \\
    \bottomrule
  \end{tabular}}%
\end{table}

\noindent\textbf{Case~1: Mirage phenomenon.}
Qwen3-VL-8B answers \emph{leelin} at both \texttt{S1-d0} and \texttt{S1-d80k}, but the source field degrades from the correct artifact to NONE. At \texttt{S1-d80k}, VC therefore remains 1 while SC and GC fall to 0, making this an outcome-preserving mirage counted by MI and reflected in OG. Outcome-only evaluation scores both states identically; only source-aware evaluation detects the grounding failure. At \texttt{S2}, the model recovers the correct source, consistent with the partial S2 recovery in Figure~\ref{fig:alluvial}.

\noindent\textbf{Case~2: Cm repair boundary.}
Under \texttt{C0} at \texttt{S1-d80k}, Qwen3-VL-30B fails to locate any source via $\Rcontext$. Under \texttt{Cm}, the retrieval-pressure instruction triggers \texttt{artifact\_recall} ($\Rtool$), which returns the correct artifact and restores GC to 1. With the model, query, and gold evidence fixed, this contrast isolates the benefit of successful tool-mediated retrieval. At \texttt{S2}, the same tool call executes but returns empty content, separating tool compliance from evidence recoverability: compaction has degraded the stored evidence, and GC falls back to 0. This encapsulates the RQ3 finding: retrieval pressure repairs pre-compaction provenance failure but cannot recover evidence that compaction has already destroyed.

%%% ============================================================
\section{Conclusion}\label{sec:conclusion}
%%% ============================================================

We presented MIRAGE, an empirical protocol for historical evidence use in multimodal personal agents under conversation-state variation.
Across seven backbones, we find that (RQ1)~pre-compaction depth and post-compaction continuation are distinct failure regimes; (RQ2)~open models are reluctant to spontaneously adopt tool-mediated retrieval even when context-based access has failed; and (RQ3)~among tool-compliant models, explicit retrieval pressure recovers provenance fidelity where original evidence remains intact, but does not reliably repair the post-compaction state produced by the current runtime.
These findings suggest that historical evidence use should be evaluated directly under state variation, with explicit attention to retrieval mechanism and provenance fidelity, rather than inferred from outcome-only correctness.

\clearpage
\begin{acks}
This research is supported by the National Key R\&D Program of China (No.\ 2023YFC3303800).
\end{acks}

% \ifshowrevisions\else\clearpage\fi
\bibliographystyle{ACM-Reference-Format}
\bibliography{references}

@inproceedings{park2023generativeagents,
  title={Generative Agents: Interactive Simulacra of Human Behavior},
  author={Park, Joon Sung and O'Brien, Joseph C. and Cai, Carrie Jun and Morris, Meredith Ringel and Liang, Percy and Bernstein, Michael S.},
  booktitle={Proceedings of the 36th Annual ACM Symposium on User Interface Software and Technology},
  articleno={2},
  numpages={22},
  year={2023},
  publisher={Association for Computing Machinery},
  address={New York, NY, USA},
  doi={10.1145/3586183.3606763},
  url={https://doi.org/10.1145/3586183.3606763}
}

@inproceedings{maharana2024locomo,
  title={Evaluating Very Long-Term Conversational Memory of LLM Agents},
  author={Maharana, Adyasha and Lee, Dong-Ho and Tulyakov, Sergey and Bansal, Mohit and Barbieri, Francesco and Fang, Yuwei},
  booktitle={Proceedings of the 62nd Annual Meeting of the Association for Computational Linguistics (Volume 1: Long Papers)},
  pages={13851--13870},
  year={2024}
}

@article{wu2025longmemeval,
  title={LongMemEval: Benchmarking Chat Assistants on Long-Term Interactive Memory},
  author={Wu, Di and Wang, Hongwei and Yu, Wenhao and Zhang, Yuwei and Chang, Kai-Wei and Yu, Dong},
  journal={arXiv preprint arXiv:2410.10813},
  year={2024}
}

@misc{memoryarena2026,
  title={MemoryArena: Benchmarking Agent Memory in Interdependent Multi-Session Agentic Tasks},
  author={He, Zexue and Wang, Yu and Zhi, Churan and Hu, Yuanzhe and Chen, Tzu-Ping and Yin, Lang and Chen, Ze and Wu, Tong Arthur and Ouyang, Siru and Wang, Zihan and others},
  journal={arXiv preprint arXiv:2602.16313},
  year={2026}
}

@inproceedings{membench2025,
  title={Membench: Towards more comprehensive evaluation on the memory of llm-based agents},
  author={Tan, Haoran and Zhang, Zeyu and Ma, Chen and Chen, Xu and Dai, Quanyu and Dong, Zhenhua},
  booktitle={Findings of the Association for Computational Linguistics: ACL 2025},
  pages={19336--19352},
  year={2025}
}

@inproceedings{cheng2024seeclick,
  title={SeeClick: Harnessing GUI Grounding for Advanced Visual GUI Agents},
  author={Cheng, Kanzhi and Sun, Qiushi and Chu, Yougang and Xu, Fangzhi and YanTao, Li and Zhang, Jianbing and Wu, Zhiyong},
  booktitle={Proceedings of the 62nd Annual Meeting of the Association for Computational Linguistics (Volume 1: Long Papers)},
  pages={9313--9332},
  year={2024}
}

@inproceedings{masry-etal-2022-chartqa,
  title={ChartQA: A Benchmark for Question Answering about Charts with Visual and Logical Reasoning},
  author={Masry, Ahmed and Long, Do Xuan and Tan, Jia Qing and Joty, Shafiq and Hoque, Enamul},
  booktitle={Findings of the Association for Computational Linguistics: ACL 2022},
  pages={2263--2279},
  year={2022},
  address={Dublin, Ireland},
  publisher={Association for Computational Linguistics},
  doi={10.18653/v1/2022.findings-acl.177},
  url={https://aclanthology.org/2022.findings-acl.177/}
}

@inproceedings{memgallery2026,
  author    = {Bei, Yuanchen and Wei, Tianxin and Ning, Xuying and Zhao, Yanjun and Liu, Zhining and Lin, Xiao and Zhu, Yada and Hamann, Hendrik and He, Jingrui and Tong, Hanghang},
  title     = {Mem-Gallery: Benchmarking Multimodal Long-Term Conversational Memory for {MLLM} Agents},
  booktitle = {Proceedings of the 64th Annual Meeting of the Association for Computational Linguistics (Volume 1: Long Papers)},
  year      = {2026}
}

@article{liu2024lost,
  title={Lost in the Middle: How Language Models Use Long Contexts},
  author={Liu, Nelson F and Lin, Kevin and Hewitt, John and Paranjape, Ashwin and Bevilacqua, Michele and Petroni, Fabio and Liang, Percy},
  journal={Transactions of the Association for Computational Linguistics},
  volume={12},
  pages={157--173},
  year={2024}
}

@article{webagent2025longcontext,
  title={Evaluating Long-Context Reasoning in LLM-Based WebAgents},
  author={Chung, Andy and Zhang, Yichi and Lin, Kaixiang and Rawal, Aditya and Gao, Qiaozi and Chai, Joyce},
  journal={arXiv preprint arXiv:2512.04307},
  year={2025}
}

@article{halumem2025,
  title={Halumem: Evaluating Hallucinations in Memory Systems of Agents},
  author={Chen, Ding and Niu, Simin and Li, Kehang and Liu, Peng and Zheng, Xiangping and Tang, Bo and Li, Xinchi and Xiong, Feiyu and Li, Zhiyu},
  journal={arXiv preprint arXiv:2511.03506},
  year={2025}
}

@inproceedings{contextlength2025emnlp,
  author    = {Du, Yufeng and Tian, Minyang and Ronanki, Srikanth and Rongali, Subendhu and Bodapati, Sravan Babu and Galstyan, Aram and Wells, Azton and Schwartz, Roy and Huerta, Eliu A and Peng, Hao},
  title     = {Context Length Alone Hurts {LLM} Performance Despite Perfect Retrieval},
  booktitle = {Findings of the Association for Computational Linguistics: EMNLP 2025},
  year      = {2025}
}

@article{hallucination_survey2025,
  title={LLM-based Agents Suffer from Hallucinations: A Survey of Taxonomy, Methods, and Directions},
  author={Lin, Xixun and Ning, Yucheng and Zhang, Jingwen and Dong, Yan and Liu, Yilong and Wu, Yongxuan and Qi, Xiaohua and Sun, Nan and Shang, Yanmin and Wang, Kun and others},
  journal={arXiv preprint arXiv:2509.18970},
  year={2025},
  eprint={2509.18970},
  archivePrefix={arXiv},
  primaryClass={cs.AI},
  url={https://arxiv.org/abs/2509.18970}
}

@article{long2026m3agent,
  title={Seeing, Listening, Remembering, and Reasoning: A Multimodal Agent with Long-Term Memory},
  author={Long, Lin and He, Yichen and Ye, Wentao and Pan, Yiyuan and Lin, Yuan and Li, Hang and Zhao, Junbo and Li, Wei},
  journal={arXiv preprint arXiv:2508.09736},
  year={2025}
}

@article{mei2026atmbench,
  title={According to Me: Long-Term Personalized Referential Memory QA},
  author={Mei, Jingbiao and Chen, Jinghong and Yang, Guangyu and Hou, Xinyu and Li, Margaret and Byrne, Bill},
  journal={arXiv preprint arXiv:2603.01990},
  year={2026}
}

@article{ma2025proofofuse,
  title={Proof-of-Use: Mitigating Tool-Call Hacking in Deep Research Agents},
  author={Ma, Shengjie and Deng, Chenlong and Mao, Jiaxin and Huang, Jiadeng and Wang, Teng and Wu, Junjie and Zhang, Changwang and Wang, Jun},
  journal={arXiv preprint arXiv:2510.10931},
  year={2025},
  eprint={2510.10931},
  archivePrefix={arXiv},
  primaryClass={cs.AI},
  url={https://arxiv.org/abs/2510.10931}
}

@misc{openclaw2026,
  title={OpenClaw: Personal AI Assistant},
  author={{OpenClaw Contributors}},
  year={2026},
  howpublished={\url{https://github.com/openclaw/openclaw}}
}

@misc{openai2025gpt5,
  title={Introducing GPT-5 for developers},
  author={{OpenAI}},
  year={2025},
  howpublished={\url{https://openai.com/index/introducing-gpt-5-for-developers}}
}

@misc{bai2025qwen3vl,
  title={Qwen3-VL Technical Report},
  author={Bai, Shuai and Cai, Yuxuan and Chen, Ruizhe and Chen, Keqin and Chen, Xionghui and Cheng, Zesen and Deng, Lianghao and Ding, Wei and Gao, Chang and Ge, Chunjiang and others},
  journal={arXiv preprint arXiv:2511.21631},
  year={2025}
}

@article{packer2023memgpt,
  title={MemGPT: Towards LLMs as Operating Systems},
  author={Packer, Charles and Wooders, Sarah and Lin, Kevin and Fang, Vivian and Patil, Shishir G and Stoica, Ion and Gonzalez, Joseph E},
  journal={arXiv preprint arXiv:2310.08560},
  year={2023}
}

@inproceedings{asai2024selfrag,
  author    = {Asai, Akari and Wu, Zeqiu and Wang, Yizhong and Sil, Avirup and Hajishirzi, Hannaneh},
  title     = {{Self-RAG}: Learning to Retrieve, Generate, and Critique through Self-Reflection},
  booktitle = {The Twelfth International Conference on Learning Representations},
  year      = {2024}
}

@article{hsieh2024ruler,
  title={RULER: What's the Real Context Size of Your Long-Context Language Models?},
  author={Hsieh, Cheng-Ping and Sun, Simeng and Kriman, Samuel and Acharya, Shantanu and Rekesh, Dima and Jia, Fei and Zhang, Yang and Ginsburg, Boris},
  journal={arXiv preprint arXiv:2404.06654},
  year={2024}
}

@article{gao2023ragsurvey,
  title={Retrieval-Augmented Generation for Large Language Models: A Survey},
  author={Gao, Yunfan and Xiong, Yun and Gao, Xinyu and Jia, Kangxiang and Pan, Jinliu and Bi, Yuxi and Dai, Yi and Sun, Jiawei and Wang, Meng and Wang, Haofen},
  journal={arXiv preprint arXiv:2312.10997},
  year={2023},
  eprint={2312.10997},
  archivePrefix={arXiv},
  primaryClass={cs.CL},
  url={https://arxiv.org/abs/2312.10997}
}

@article{huang2025hallucination_survey,
  title={A Survey on Hallucination in Large Language Models: Principles, Taxonomy, Challenges, and Open Questions},
  author={Huang, Lei and Yu, Weijiang and Ma, Weitao and Zhong, Weihong and Feng, Zhangyin and Wang, Haotian and Chen, Qianglong and Peng, Weihua and Feng, Xiaocheng and Qin, Bing and others},
  journal={ACM Transactions on Information Systems},
  volume={43},
  number={2},
  pages={1--55},
  year={2025},
  publisher={ACM New York, NY}
}

@inproceedings{shi2023irrelevant,
  title={Large Language Models Can Be Easily Distracted by Irrelevant Context},
  author={Shi, Freda and Chen, Xinyun and Misra, Kanishka and Scales, Nathan and Dohan, David and Chi, Ed H. and Sch{\"a}rli, Nathanael and Zhou, Denny},
  booktitle={Proceedings of the 40th International Conference on Machine Learning},
  pages={31210--31227},
  year={2023},
  volume={202},
  series={Proceedings of Machine Learning Research},
  publisher={PMLR},
  address={Honolulu, Hawaii, USA},
  url={https://proceedings.mlr.press/v202/shi23a.html}
}

@inproceedings{zhang2024infinitybench,
  title={$\infty${B}ench: Extending Long Context Evaluation Beyond 100{K} Tokens},
  author={Zhang, Xinrong and Chen, Yingfa and Hu, Shengding and Xu, Zihang and Chen, Junhao and Hao, Moo Khai and Han, Xu and Thai, Zhen Leng and Wang, Shuo and Liu, Zhiyuan and Sun, Maosong},
  booktitle={Proceedings of the 62nd Annual Meeting of the Association for Computational Linguistics (Volume 1: Long Papers)},
  pages={15262--15277},
  year={2024},
  address={Bangkok, Thailand},
  publisher={Association for Computational Linguistics},
  doi={10.18653/v1/2024.acl-long.814},
  url={https://aclanthology.org/2024.acl-long.814/}
}

@article{shinn2023reflexion,
  title={Reflexion: Language Agents with Verbal Reinforcement Learning},
  author={Shinn, Noah and Cassano, Federico and Gopinath, Ashwin and Narasimhan, Karthik and Yao, Shunyu},
  journal={Advances in neural information processing systems},
  volume={36},
  pages={8634--8652},
  year={2023}
}

@inproceedings{kamradt2023needle,
  title={Needle-in-the-Haystack Testing LLMs with a Complex Reasoning Task},
  author={Schuster, Thomas and Lambert, Marian and D{\"o}ring, Nico and Tr{\"o}gele, Julius},
  booktitle={International Conference on Engineering Applications of Neural Networks},
  pages={254--266},
  year={2025},
  organization={Springer}
}

@article{wang2025internvl35,
  title={{InternVL3.5}: Advancing Open-Source Multimodal Models in Versatility, Reasoning, and Efficiency},
  author={Wang, Weiyun and Gao, Zhangwei and Gu, Lixin and Pu, Hengjun and Cui, Long and Wei, Xingguang and Liu, Zhaoyang and Jing, Linglin and Ye, Shenglong and Shao, Jie and others},
  journal={arXiv preprint arXiv:2508.18265},
  year={2025},
  eprint={2508.18265},
  archivePrefix={arXiv},
  primaryClass={cs.CV},
  url={https://arxiv.org/abs/2508.18265}
}

@article{gemma2025gemma3,
  title={{Gemma 3 Technical Report}},
  author={{Gemma Team, Google DeepMind}},
  journal={arXiv preprint arXiv:2503.19786},
  year={2025},
  eprint={2503.19786},
  archivePrefix={arXiv},
  primaryClass={cs.CL},
  url={https://arxiv.org/abs/2503.19786}
}

@misc{anthropic2025claudehaiku,
  author       = {{Anthropic}},
  title        = {Introducing Claude Haiku 4.5},
  howpublished = {\url{https://www.anthropic.com/news/claude-haiku-4-5}},
  year         = {2025}
}

@article{zhang2025enhancing,
  author  = {Zhang, Wenxiao and Kong, Xiangrui and Dewitt, Conan and Br{\"a}unl, Thomas and Hong, Jin B.},
  title   = {Enhancing reliability in {LLM}-integrated robotic systems: A unified approach to security and safety},
  journal = {Journal of Systems and Software},
  year    = {2026}
}

@article{liu2026prisma,
  title={{PRISMA}: Reinforcement Learning Guided Two-Stage Policy Optimization in Multi-Agent Architecture for Open-Domain Multi-Hop Question Answering},
  author={Liu, Yu and Zhang, Wenxiao and Cao, Cong and Lu, Wenxuan and Yuan, Fangfang and Guo, Diandian and Peng, Kun and Sun, Qiang and Zhang, Kaiyan and Liu, Yanbing and others},
  journal={arXiv preprint arXiv:2601.05465},
  year={2026},
  eprint={2601.05465},
  archivePrefix={arXiv},
  primaryClass={cs.AI},
  url={https://arxiv.org/abs/2601.05465}
}

@inproceedings{long2025revisiting,
  title={Revisiting multimodal fusion for 3D anomaly detection from an architectural perspective},
  author={Long, Kaifang and Xie, Guoyang and Ma, Lianbo and Liu, Jiaqi and Lu, Zhichao},
  booktitle={Proceedings of the AAAI Conference on Artificial Intelligence},
  volume={39},
  number={12},
  pages={12273--12281},
  year={2025}
}

@misc{zhao2026seeingendstepzero,
  title={Seeing the End at Step Zero: Accelerating Diffusion MLLMs via MLP Sparsity-Aware Truncation},
  author={Zhao, Qicheng and Sun, Qi and Yan, Zheyu},
  year={2026},
  eprint={2607.14557},
  archivePrefix={arXiv},
  primaryClass={cs.AI},
  url={https://arxiv.org/abs/2607.14557}
}

@misc{wang2026magesafeguardingllmagents,
  title={MAGE: Safeguarding LLM Agents against Long-Horizon Threats via Shadow Memory},
  author={Wang, Yuhui and Jiang, Tanqiu and Liang, Jiacheng and Fleming, Charles and Wang, Ting},
  year={2026},
  eprint={2605.03228},
  archivePrefix={arXiv},
  primaryClass={cs.CR},
  url={https://arxiv.org/abs/2605.03228}
}

@inproceedings{liu2026llms,
  title={Do {LLM}s Know Tool Irrelevance? Demystifying Structural Alignment Bias in Tool Invocations},
  author={Liu, Yilong and Lin, Xixun and Cao, Pengfei and Zhang, Ge and Fang, Fang and Cao, Yanan},
  booktitle={Proceedings of the 64th Annual Meeting of the Association for Computational Linguistics (Volume 1: Long Papers)},
  pages={31934--31958},
  year={2026}
}

@inproceedings{li2026revisor,
  title={Revisor: Beyond Textual Reflection, towards Multimodal Introspective Reasoning in Long-Form Video Understanding},
  author={Li, Jiaze and Yin, Hao and Tan, Wenhui and Chen, Jingyang and Xu, Boshen and Qu, Yuxun and Chen, Yijing and Ju, Jianzhong and Luo, Zhenbo and Luan, Jian},
  booktitle={Proceedings of the IEEE/CVF Conference on Computer Vision and Pattern Recognition},
  pages={5059--5069},
  year={2026}
}

@article{lin2026safeharness,
  title={SafeHarness: Lifecycle-Integrated Security Architecture for {LLM}-based Agent Deployment},
  author={Lin, Xixun and Liu, Yang and Chen, Yancheng and Wu, Yongxuan and Ning, Yucheng and Liu, Yilong and Sun, Nan and Zhang, Shun and Chong, Bin and Zhou, Chuan and Cao, Yanan},
  journal={arXiv preprint arXiv:2604.13630},
  year={2026},
  eprint={2604.13630},
  archivePrefix={arXiv},
  primaryClass={cs.CR},
  url={https://arxiv.org/abs/2604.13630}
}

@misc{wu2026promsa,
  title={{ProMSA}: Progressive Multimodal Search Agents for Knowledge-Based Visual Question Answering},
  author={Wu, ZhengXian and Xu, Hangrui and Shi, Kai and Chen, Zhuohong and Yu, Yunyao and Zhang, Chuanrui and Liao, Zirui and Yang, Jun and Yang, Zhenyu and Lu, Haonan and Wang, Haoqian},
  year={2026},
  eprint={2606.27974},
  archivePrefix={arXiv},
  primaryClass={cs.CV},
  url={https://arxiv.org/abs/2606.27974}
}

@article{zhang2025cogmem,
  title={{CogMem: A Cognitive Memory Architecture for Sustained Multi-Turn Reasoning in Large Language Models}},
  author={Zhang, Yiran and Hu, Jincheng and Dras, Mark and Naseem, Usman},
  journal={arXiv preprint arXiv:2512.14118},
  year={2025},
  eprint={2512.14118},
  archivePrefix={arXiv},
  primaryClass={cs.CL},
  url={https://arxiv.org/abs/2512.14118}
}

\clearpage
\appendix
\section{Appendices}\label{app:supplementary}

\subsection{Evidence-Type Extension}\label{app:evidence_extension}

We apply the same protocol and deterministic scoring to 109 additional probes (55 answerable and 54 unanswerable) over six evidence types: a natural photo, infographic, code file, email, invoice PDF, and report-with-table PDF.
Table~\ref{tab:app_evidence_extension} compares source correctness (SC) on this extension with the original chart/UI evidence set under \texttt{C0}.
The extension is harder in absolute terms for the open VLMs, but all three Qwen sizes again show substantially lower SC at \texttt{S1-d80k} and \texttt{S2} than at \texttt{S1-d0}.
Thus, the state-conditioned source-recovery decline is not confined to charts and user-interface artifacts.

\begin{table}[H]
  \centering
  \small
  \setlength{\tabcolsep}{2pt}
  \renewcommand{\arraystretch}{0.9}
  \caption{SC across the original chart/UI evidence and six additional evidence types under \texttt{C0}. The original set has 100 answerable probes and the extension has 55; values are percentages.}
  \label{tab:app_evidence_extension}
  \begin{tabularx}{\columnwidth}{@{}llYYYY@{}}
    \toprule
    Model & Evidence set & \texttt{S1-d0} & \texttt{S1-d50k} & \texttt{S1-d80k} & \texttt{S2} \\
    \midrule
    \multirow{2}{*}{Qwen3-VL-4B} & Original & 71 & 32 & 3 & 52 \\
                                  & Extension & 46 & 14 & 16 & 23 \\
    \midrule
    \multirow{2}{*}{Qwen3-VL-8B} & Original & 90 & 78 & 27 & 79 \\
                                  & Extension & 51 & 33 & 15 & 35 \\
    \midrule
    \multirow{2}{*}{Qwen3-VL-30B} & Original & 82 & 72 & 47 & 53 \\
                                   & Extension & 39 & 23 & 12 & 25 \\
    \bottomrule
  \end{tabularx}
\end{table}

\subsection{Compaction-Strategy Ablation}\label{app:compaction_ablation}

At \texttt{S2}, we hold the evidence, questions, Qwen answerer, and scorer fixed while varying only how GPT-5 writes the compaction summary.
The verbatim condition preserves the paper's default summary, the free-form condition rewrites the same paths and values, and the per fact citations condition attaches a source path to each value.
Table~\ref{tab:app_compaction_ablation} reports SC/GC for all three Qwen sizes.
Per fact citations substantially raise SC for Qwen3-VL-8B (79 to 95) while GC remains near 45; GC decreases for Qwen3-VL-4B and Qwen3-VL-30B.
Summary format can therefore improve attribution without reliably improving grounding in this stack.

\begin{table}[H]
  \centering
  \small
  \setlength{\tabcolsep}{2pt}
  \renewcommand{\arraystretch}{0.9}
  \caption{SC/GC at \texttt{S2} under \texttt{C0} for three GPT-5 summary formats. Verbatim is the paper default; each cell reports percentages over 100 answerable probes.}
  \label{tab:app_compaction_ablation}
  \begin{tabularx}{\columnwidth}{@{}lYYY@{}}
    \toprule
    Answerer & Verbatim & Free form & Per fact citations \\
    \midrule
    Qwen3-VL-4B  & 52 / 34 & 63 / 22 & 64 / 24 \\
    Qwen3-VL-8B  & 79 / 44 & 94 / 44 & 95 / 47 \\
    Qwen3-VL-30B & 53 / 38 & 48 / 31 & 52 / 34 \\
    \bottomrule
  \end{tabularx}
\end{table}

\subsection{Question Construction and Filler Controls}\label{app:data_controls}

The question bank contains 200 manually constructed probes over six planted evidence objects: 100 answerable questions and 100 matched unanswerable counterparts, organized into 12 object--answerability families.
Answerable probes target concrete values rendered in each chart or UI artifact; their unanswerable counterparts preserve the topic, local cues, and response format while removing supporting evidence.
Each annotation records the gold answerability judgment, canonical source path, normalized value, difficulty (lookup, comparison/normalization, or layout-sensitive extraction), query type, and shortcut risk from priors, filenames, templates, or lexical cues.

Filler turns are sampled with a fixed seed from an independently authored library of 20 persona-consistent software-engineering and machine-learning tasks.
They are deliberately topic-disjoint from planted evidence and contain no target answers, artifact identifiers, or topic-matched distractors.
Their sole purpose is to increase input depth and trigger native compaction while leaving the planted evidence unchanged.

\subsection{Tool-Call Compliance Check}\label{app:tool_compliance}

Tool-call parseability is determined jointly by a backbone's emitted format and the corresponding standard per-model parser integrated by the runtime.
Qwen3-VL emits Hermes-style tool-call blocks accepted by its matching parser, whereas InternVL3.5-20B and Gemma-3-27B emit inconsistent formats under the current orchestration contract.
As a controlled check, a minimal 30-line adapter maps one VLM's native tool tokens into the expected parser format without changing its output content.
For the four inspected cases, the adapter changes the recorded access path from \texttt{R\_context} to \texttt{R\_tool} in all 4/4 cases, confirming that the diagnostic distinguishes model-side formatting compliance from the content of the returned answer.

\subsection{Benchmark and Depth Context}\label{app:benchmark_depth}

Table~\ref{tab:app_benchmark_gap} contrasts shared-backbone scores from ATM-Bench~\cite{mei2026atmbench} with MIRAGE's state-conditioned diagnostics.
Table~\ref{tab:app_benchmark_axes} summarizes the conceptual distinction from ATM-Bench and MemoryArena~\cite{memoryarena2026}: MIRAGE manipulates conversation state while holding planted evidence, questions, and scoring fixed, and separates source attribution from grounded correctness.

\begin{table}[H]
  \centering
  \small
  \setlength{\tabcolsep}{2pt}
  \renewcommand{\arraystretch}{0.9}
  \caption{Existing-benchmark scores versus MIRAGE's state-conditioned diagnostic gap for shared backbones.}
  \label{tab:app_benchmark_gap}
  \begin{tabularx}{\columnwidth}{@{}ll>{\raggedright\arraybackslash}X@{}}
    \toprule
    Backbone & ATM-Bench oracle & MIRAGE under state variation \\
    \midrule
    Qwen3-VL-8B & QS 77.8 (Hard 47.3) & \texttt{S1-d80k}: joint-event MI 0.78; separate OG 0.29 \\
    GPT-5 & QS 85.3 (Hard 74.7) & MI $\leq 0.11$ and OG $\leq 0.01$ across states \\
    \bottomrule
  \end{tabularx}
\end{table}

\begin{table}[H]
  \centering
  \small
  \setlength{\tabcolsep}{2pt}
  \renewcommand{\arraystretch}{0.9}
  \caption{Conceptual comparison with the closest memory benchmarks.}
  \label{tab:app_benchmark_axes}
  \begin{tabularx}{\columnwidth}{@{}>{\raggedright\arraybackslash}p{0.46\columnwidth}YYY@{}}
    \toprule
    Axis & \shortstack{Memory\\Arena} & ATM-Bench & MIRAGE \\
    \midrule
    Conversation state as sole controlled variable & no & no & \textbf{yes} \\
    Per-claim source attribution (SC) & no & partial & \textbf{yes} \\
    Pre/post-compaction regime separation & no & no & \textbf{yes} \\
    Planted multimodal evidence & no & partial & \textbf{yes} \\
    Diagnoses mechanism vs. outcome only & partial & partial & \textbf{yes} \\
    \bottomrule
  \end{tabularx}
\end{table}

The fixed 50k/80k checkpoints are lifecycle probes rather than tuned thresholds.
Table~\ref{tab:app_depth_normalization} reports the fraction of each advertised context window represented by \texttt{d80k} and the associated depth sensitivity and retention.
The same absolute coordinate spans 20\% of GPT-5's window but 63\% of Gemma-3's, so open models are not advantaged by the fixed checkpoints.
Within Qwen3-VL, where the context window and architecture family are controlled, larger models show lower depth sensitivity and higher retention.

\begin{table}[H]
  \centering
  \small
  \setlength{\tabcolsep}{2pt}
  \renewcommand{\arraystretch}{0.9}
  \caption{Depth normalization and within-family size trend.}
  \label{tab:app_depth_normalization}
  \begin{tabularx}{\columnwidth}{@{}lYYYY@{}}
    \toprule
    Backbone/family & Window & \texttt{d80k} & DS & Ret \\
    \midrule
    GPT-5 & 400K & 20\% & 3.4 & 94.0 \\
    Claude-4.5-Haiku & 200K & 40\% & 11.8 & 73.9 \\
    Gemma-3-27B & 128K & 63\% & 48.6 & 42.1 \\
    Qwen3-VL-4B & 256K & 31\% & 98.1 & 1.9 \\
    Qwen3-VL-8B & 256K & 31\% & 74.2 & 21.7 \\
    Qwen3-VL-30B & 256K & 31\% & 43.1 & 51.4 \\
    \bottomrule
  \end{tabularx}
\end{table}

\subsection{Statistical Uncertainty and Routing Stability}\label{app:uncertainty}

Table~\ref{tab:app_wilson_ci} reports Wilson 95\% confidence intervals using the observed valid-probe denominator for each representative single-run cell.
These intervals quantify within-run binomial uncertainty rather than variability across the three runs; the reported main-table metrics remain averages over three independent runs.
The intervals preserve the main contrasts, including the separation between Qwen3-VL-4B's GC collapse and Qwen3-VL-30B's retained GC at \texttt{S1-d80k}.

\begin{table}[H]
  \centering
  \small
  \setlength{\tabcolsep}{2pt}
  \renewcommand{\arraystretch}{0.87}
  \caption{Representative single-run Wilson 95\% confidence intervals.}
  \label{tab:app_wilson_ci}
  \begin{tabularx}{\columnwidth}{@{}>{\raggedright\arraybackslash}p{0.48\columnwidth}YYY@{}}
    \toprule
    Cell & Count & Rate & 95\% CI \\
    \midrule
    Qwen3-VL-30B GC, \texttt{S1-d80k} & 41/100 & 41.0\% & [31.9, 50.8] \\
    Qwen3-VL-4B GC, \texttt{S1-d80k}  & 1/100  & 1.0\%  & [0.2, 5.4] \\
    Gemma-3-27B SC, \texttt{S1-d80k} & 70/99  & 70.7\% & [61.1, 78.8] \\
    GPT-5 GC, \texttt{S2}             & 84/99  & 84.8\% & [76.5, 90.6] \\
    \bottomrule
  \end{tabularx}
\end{table}

Routing is also stable at the scale relevant to the paper's mechanism claims.
GPT-5 uses the tool route for 50/200 \texttt{S2} calls (25\%) while maintaining GC${=}$84.8\%.
Qwen3-VL-4B and Qwen3-VL-8B remain at 0/199--200 tool calls across all states, whereas Qwen3-VL-30B uses tools only at \texttt{S1-d0} (34/200) and \texttt{S2} (28/200).
Thus, GPT-5's stability reflects a different route mixture rather than context continuity alone.

\end{document}